\documentclass[11pt]{article}
\usepackage[preprint]{acl}
\usepackage{latexsym}
\usepackage{amsmath}
\usepackage{amsfonts}
\usepackage{amssymb}
\usepackage{algorithm}
\usepackage{algorithmic}
\usepackage{subcaption}
\usepackage{microtype}
\usepackage{graphicx}
\usepackage{float}
\usepackage{placeins}
\usepackage{booktabs}
\usepackage{multirow}
\usepackage{url}
\usepackage[T1]{fontenc}
\usepackage{times}

\newcommand{\STRRepo}{\url{https://github.com/Phoenix-ni/STR.git}}

\title{Semantic Triplet Restoration: A Novel Protocol for Hierarchical Table Understanding in Large Language Models}

\author{
  \textbf{Yibin Zhao}$^{a,*}$, \textbf{Fangxin Shang}$^{b}$, \textbf{Dingrui Yang}$^{a}$, \textbf{Yuqi Wang}$^{c,*}$ \\
  $^{a}$Taiyuan University of Technology \\
  $^{b}$AI Lab, Qifu Technology, Beijing, China \\
  $^{c}$AI Lab, Greensea Technology, Shenzhen, China \\
  \texttt{2024005305@link.tyut.edu.cn}, \texttt{gprofessorsfx@gmail.com}, \\
  \texttt{yangdingrui5759@link.tyut.edu.cn}, \texttt{wangyuqi@green-sea.cn} \\
  $^{*}$Corresponding authors
}

\begin{document}


\maketitle

\begin{abstract}
Table question answering requires models to recover semantic relations encoded implicitly by two-dimensional layout, merged cells, and hierarchical headers. Current pipelines typically use HTML or Markdown as intermediate table representations, but these layout-oriented serializations introduce markup overhead and require large language models to infer header-cell alignments from row and column spans. We propose Semantic Triplet Restoration (STR), a protocol that rewrites each cell as an atomic fact $\langle\text{item path}, \text{feature path}, \text{value}\rangle$, where the item path specifies the row-wise entity, the feature path specifies the hierarchical attribute, and the value contains the cell content. We also present TripletQL, a lightweight query-aware router that uses STR to select an appropriate rendering or filtered subset of triplets for each question. Across four Chinese and English table-QA benchmarks, STR matches or improves upon HTML-based baselines while reducing input tokens. The relative benefit grows for smaller language models and longer table contexts, suggesting that explicit semantic representations are especially useful under constrained inference budgets. Code and data are available at \STRRepo.
\end{abstract}

\section{Introduction}

\begin{figure*}[!t]
    \centering
    \includegraphics[width=0.8\textwidth]{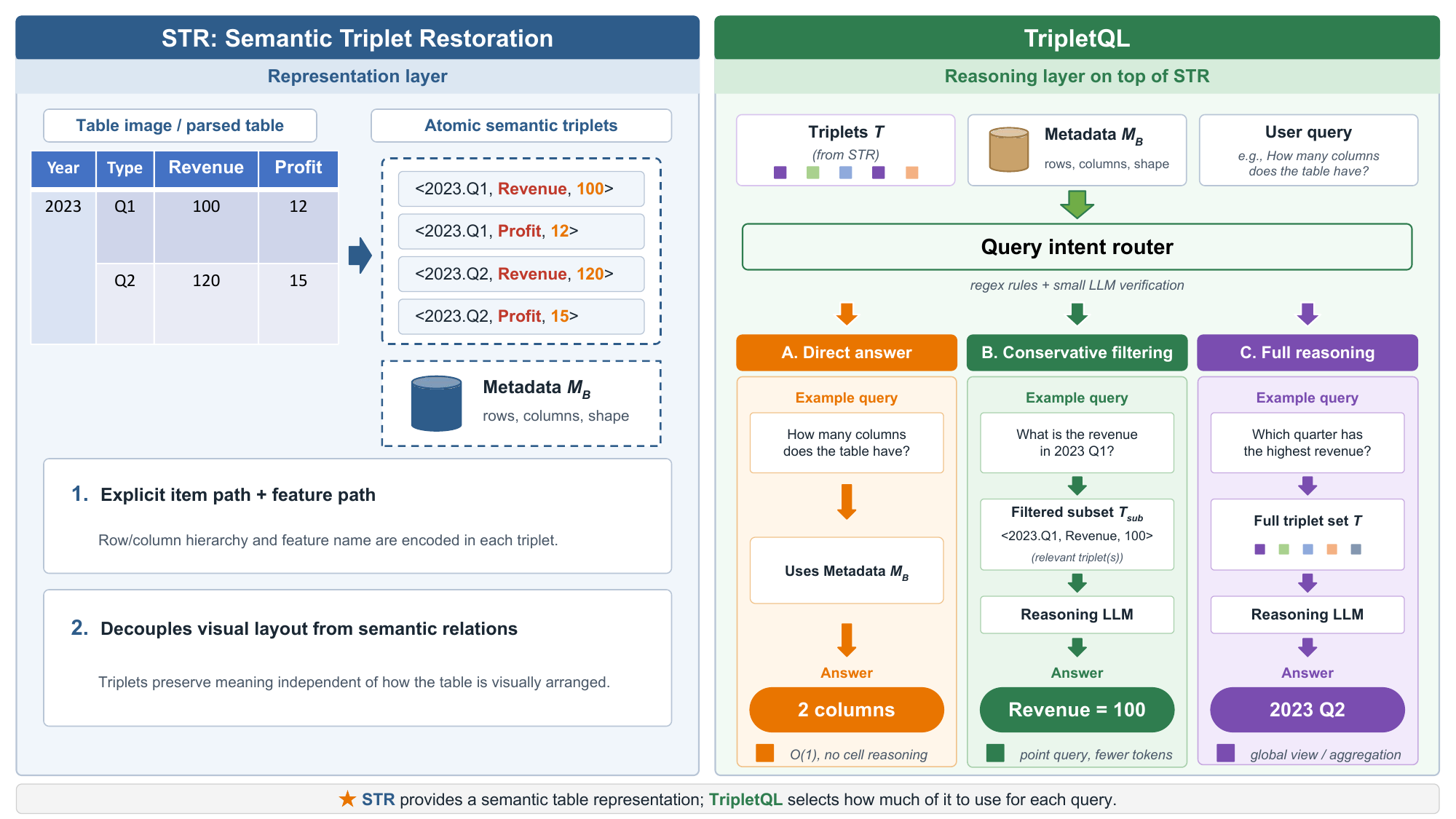}
    \caption{An example of a complex hierarchical table from the TableEval dataset.}
    \label{fig:intro_table}
\end{figure*}

Table understanding and reasoning is a core task in document intelligence \citep{chen2019tabfact}. Recent end-to-end vision-language models such as PaddleOCR-VL \citep{baidu2025paddleocrvl}, MinerU \citep{wang2024mineru}, and MonkeyOCR \citep{monkey2025monkeyocr} have made notable progress on table recognition, predominantly by restoring tables from images into HTML, which has become the de facto intermediate representation for downstream LLM-based question answering. Yet once the downstream objective shifts from \emph{structural restoration} to \emph{reasoning over the restored table}, this intermediate layer begins to expose its limitations.

HTML conveys \emph{visual layout}, but QA relies on \emph{semantic relations}. In HTML, these are coupled, forcing LLMs to reconstruct 2D grids and header alignments before reasoning. This coupling inflates context with weak layout tags and risks numerical drift when LLMs miscount cross-row/column spans \citep{sui2024table, li2024llm}.

We propose \emph{Semantic Triplet Restoration} (STR), decoupling layout and semantics. STR rewrites each cell as an atomic fact $\langle\text{item path}, \text{feature path}, \text{value}\rangle$, transforming headers into explicit, dot-concatenated semantic paths (e.g., $\langle 2023.Q1, \text{Revenue}, 100 \rangle$). Visual layout is confined to parsing, presenting the LLM with a semantic view that avoids in-context grid reconstruction. To demonstrate STR's utility for query-aware context selection, we introduce TripletQL, a lightweight router that filters triplets by intent and mitigates long-context bias via multi-turn context re-injection \citep{liu2024lost}.

We systematically evaluate STR on four complementary table-QA benchmarks spanning Chinese and English: TableEval-test \citep{zhu2025tableeval} (sparse-evidence retrieval, Chinese), WikiTableQuestions \citep{pasupat2015wtq} (global aggregation, English), TableBench \citep{wu2025tablebench} (integrated analysis), and TQA-Bench \citep{qiu2024tqabench} (multi-table long-context). Under matched parameter and token budgets, our method preserves or improves accuracy while consistently reducing token cost. Moreover, across three backbone scales (GLM-4.5-Air $\to$ LongCat-Lite $\to$ Qwen3-0.6B) we observe that the relative gain from STR rises monotonically, indicating that the representation is especially valuable for resource-constrained deployments. If agentic systems become the default interface for table-heavy workloads, token budget will act less like a secondary engineering detail and more like a direct constraint on throughput and cost.

Our contributions are as follows:
\begin{itemize}
    \item We propose STR, a representation-level decoupling protocol whose unit is the $\langle\text{item path}, \text{feature path}, \text{value}\rangle$ triplet, restoring hierarchical headers from visual alignment into explicit semantic paths (\S\ref{sec:proposed}).
    \item On four heterogeneous benchmarks, we systematically verify STR's advantage over the HTML paradigm: it preserves or improves accuracy while substantially reducing input tokens, with the relative gain rising monotonically as the backbone shrinks, making the protocol especially valuable for resource-constrained deployments (\S\ref{sec:qa_performance}, \S\ref{sec:model_tiers}).
    \item We build the TripletQL Agent on top of STR as a lightweight routing instance: feature-triggered routing with regex plus selective small-model confirmation, conservative filtering, and multi-turn attention engineering to mitigate lost-in-the-middle (\S\ref{sec:routing_gate}, \S\ref{sec:conservative_extraction}).
\end{itemize}

\section{Related Work}

\subsection{Visual Table Structure Recognition}

Table Structure Recognition (TSR) aims to recover row/column boundaries, merged cells, and multi-level headers from images or PDFs. Early work followed a cell-level pipeline route---exemplified by PaddleX, which explicitly separates detection, rectification, cell recognition, and HTML restoration---and matured on ruled tables. Since 2023, this line has been progressively replaced by end-to-end vision-language generation: PaddleOCR-VL \citep{baidu2025paddleocrvl}, MonkeyOCR \citep{monkey2025monkeyocr}, MinerU \citep{wang2024mineru}, and more recent InstructTable \citep{instructtable2026} and TDATR \citep{tdatr2026} unify the task into single-stage sequence generation with HTML or Markdown as the dominant output, while \citet{zhou2025enhancing} further introduce neighborhood-guided tool-chain reasoning for complex tables.

Most evaluations focus on OCR fidelity (e.g., TEDS) rather than downstream reasoning. For QA, HTML's reliance on \texttt{rowspan}/\texttt{colspan} tags forces LLMs to implicitly reconstruct header alignments, risking numerical drift on long tables. We thus benchmark PaddleOCR-VL-1.5 and MinerU to directly compare HTML vs.\ STR under realistic pipelines.

\subsection{Tabular Representation for LLMs}

Table representations for LLMs broadly fall into three families: serialised formats such as HTML / Markdown / CSV, graph structures, and tree structures. \citet{sui2024table} and \citet{li2024llm} report empirically that serialised forms incur substantial token redundancy on long or nested tables and require the LLM to implicitly rebuild the 2D grid in context. Structured-injection approaches---TaBERT \citep{yin2020tabert}, TAPAS \citep{herzig2020tapas}, TableFormer \citep{yang2022tableformer}, and rLLM \citep{rllm2024}---aim to alleviate this issue but still bridge through visual tags.

Hierarchical representations such as Tree-of-Table \citep{ji2024treeoftable} take a further step at the level of structural organisation, but tend to operate as a post-hoc reorganisation and do not directly provide atomic semantic coordinates at the cell level. Overall, existing representations either retain the visual layout or reorganise it after the fact, leaving open the question of a native protocol that operates in semantic-path units from the cell upwards---this is the gap STR aims to fill.

\subsection{LLM Agents in Table Reasoning}

LLMs exhibit strong zero/few-shot capabilities for table reasoning \citep{oses-grijalba-etal-2024-question}, and agent frameworks further introduce planning, tool use, and multi-agent collaboration \citep{tyagi-etal-2025-aestar}. These agents are mostly built on top of dense visual representations such as HTML, so on hierarchical long tables, the cost of intent classification and context compression still leaves room for further optimisation. In a concurrent line of work, \citet{sen2025grep} compare grep against vector retrieval in agentic retrieval settings and observe that ``retrieval'' is closer to ``retrieval + harness''---the relative ordering of grep and vector can reverse under different runtimes for the same retriever.

We adopt and extend this framing: measuring representation and routing separately---HTML vs.\ Full Triplet for the representation itself, and Full Triplet vs.\ TripletQL to isolate routing and filtering---and attribute the primary contribution to the representation. In addition, \citet{liu2024lost} report that LLMs tend to underuse the middle of long contexts (Lost-in-the-Middle); in multi-turn table dialogues, tables embedded in history can lead to hallucinations or history echoing, and our multi-turn context engineering (\S\ref{sec:conservative_extraction}) is designed around this attention bias.

\begin{figure*}[!t]
    \centering
    \includegraphics[width=0.8\textwidth]{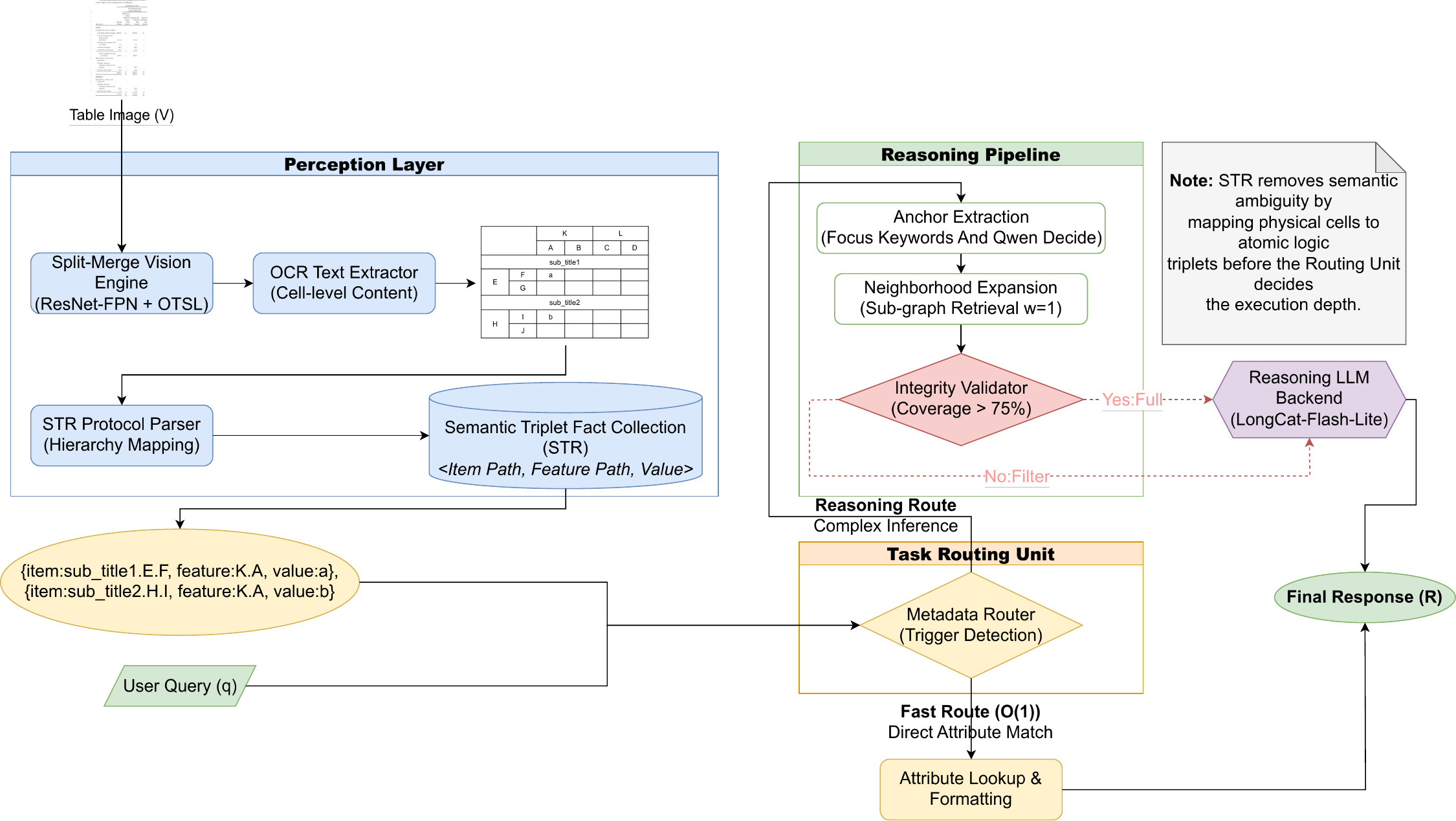}
    \caption{Processing pipeline of the TripletQL agent.}
    \label{fig:hero_pipeline}
\end{figure*}

\section{Proposed Method}
\label{sec:proposed}

Returning to the two limitations raised in \S1 (token redundancy and numerical drift), this section presents a unified solution that handles representation and reasoning at two separate layers. At the \emph{representation layer}, the STR protocol restores each cell into an atomic fact carrying an explicit semantic path, and a companion non-autoregressive visual foundation is responsible for reliably recovering this triplet structure from the image. At the \emph{reasoning layer}, TripletQL dispatches user queries into different execution paths via hierarchical feature detection, and conditionally filters or renders the context under long-context and multi-turn settings. The remainder of this section unfolds in the order ``problem formalization $\to$ STR protocol $\to$ visual foundation $\to$ reasoning agent''; the overall architecture is shown in Figure~\ref{fig:hero_pipeline}.

\subsection{Problem Formalization}

Given a table image $\mathbf{I}$ and a natural-language query $q$, existing methods typically fit $\hat{a} = \arg\max_a P(a \mid q, \text{HTML}(\mathbf{I}))$ directly, feeding $O(|\mathbf{I}|)$ HTML tags as a whole into the attention window. In contrast, we first obtain the STR state through visual parsing, $\mathcal{T} = \text{STR}(\mathbf{I})$, and route over triplets rather than HTML via a feature-routing gate $\pi_R$ that shunts the decision:
\begin{equation}
    a_{t+1} = \begin{cases}
    \pi_{R}(\cdot | q, \mathcal{M}_{B}), & q \text{ triggers fast route} \\
    \pi_{D}(\cdot | q, \mathcal{T}_{sub}), & \text{otherwise}
    \end{cases}
\end{equation}
$\mathcal{M}_{B}$ is the global metadata of the STR state $\mathcal{T}$ (row/column counts, table shape), and $\mathcal{T}_{sub}$ is the filtered subset. On path B, $\pi_D$ is instantiated by the same main reasoning model, acting first as an entity filter and then as the answerer, rather than two independent modules.

\subsection{The Semantic Triplet Representation (STR) Protocol}

Planning and reasoning need a structured underlying state. The STR protocol is what we use for that state: after visual parsing, the table image $\mathbf{I}$ is mapped to a semantic-fact set $\mathcal{T} = \text{STR}(\mathbf{I})$:
\begin{equation}
    \mathcal{T} = \{ \langle \mathcal{P}_i^{\text{item}}, \mathcal{P}_j^{\text{feat}}, v_{i,j} \rangle \}
\end{equation}
Here, $v$ is the value and $\mathcal{P}$ a dot-concatenated path. Merged cells are flattened into independent coordinates to prevent misalignment. Unlike HTML, which requires LLMs to resolve complex \texttt{rowspan} tags:
\begin{quote}
\ttfamily\small
<th rowspan="2">2023</th><th>Q1</th>\\
...<td>100</td>
\end{quote}
STR directly encodes this as $\langle 2023.Q1, \text{revenue}, 100\rangle$, providing explicit cross-level headers without inference-time reconstruction.

\subsection{Optimized Non-autoregressive Visual Foundation}

\begin{figure}[!t]
    \centering
    \includegraphics[width=\linewidth]{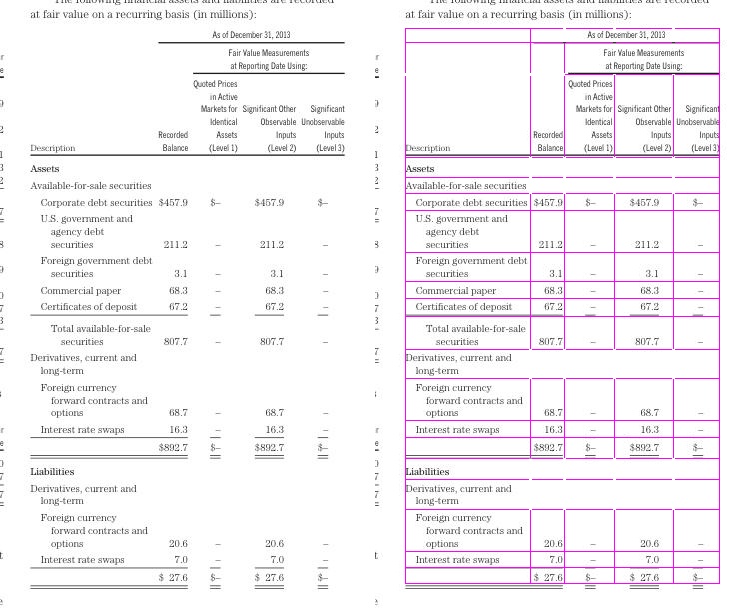}
    \caption{Visualization of visual grid prediction and cell merging.}
    \label{fig:split_merge_results}
\end{figure}

We adapt the encoder-only, non-autoregressive visual foundation of \citet{hou2025tablet}. Visual parsing is not the main novelty of this paper; its purpose is to provide STR with stable cell boundaries and hierarchical paths. The original scheme may exhibit line leakage and coordinate drift on ultra-long/ultra-wide tables, for which we redesign its loss function; representative grid prediction and merging results are shown in Figure~\ref{fig:split_merge_results}.

\noindent\textbf{Split: 1D dilated soft labels and inverse dynamic weighting (IDW).} We replace the hard label binary classification of the original scheme with soft label Focal Loss: perform spatial dilation with radius $k=1$ on pixel-level ground truth to accommodate subtle coordinate fluctuations, and adaptively allocate weights according to axial training accuracy with inverse dynamic weighting ($w_{\text{row}} = \mathcal{L}_{\text{col}}/(\mathcal{L}_{\text{row}}+\mathcal{L}_{\text{col}})$), preventing any single branch from dominating the gradient and improving the robustness of continuous merged regions across rows/columns.

\noindent\textbf{Merge: pruned OTSL classes and masked alignment.} Since explicit grid indices have been established in the Split stage, OTSL cell classification is simplified from the original 5 categories to 4 categories (C/L/U/X), removing the redundant "New Line (NL)" identifier; at the same time, an \texttt{ignore\_index} mask is introduced for the padded areas in variable-length batches to mitigate the risk of coordinate jitter in large-scale tensor operations in the original scheme. Full loss definitions and mask details are in Appendix~\ref{sec:visual_loss_detail}.

\subsection{The TripletQL Agent Architecture}

On top of STR, we design a reasoning architecture with feature-triggered routing as the entry point and conservative filtering as the deep path. It has two complementary sub-systems: hierarchical feature detection and routing for query intent, and a multi-stage conservative filtering pipeline for context compression. The full flow is shown in Figure~\ref{fig:hero_pipeline}; the simplified decision pseudocode is in Algorithm~\ref{alg:routing}; detailed referential expansion, coverage circuit-breaking, and multi-turn context re-injection are deferred to Appendix~\ref{sec:full_algorithm}.

\begin{algorithm}[H]
\caption{Simplified decision logic of TripletQL.}
\label{alg:routing}
\begin{algorithmic}[1]
\REQUIRE User instruction $q$, STR state $\mathcal{T}$, metadata $\mathcal{M}_B$
\STATE $\mathcal{F} \leftarrow \pi_R(q)$ \COMMENT{Feature-triggered routing (regex + Qwen verification)}
\IF{$\texttt{direct\_answer} \in \mathcal{F}$} \STATE Path A (Metadata direct-answer)
\ELSIF{$\operatorname{CanFilter}(q, \mathcal{F}, |\mathcal{T}|)$}
    \STATE $\mathcal{T}_{sub} \leftarrow \operatorname{ConservativeExtract}(q, \mathcal{T})$ \COMMENT{Path B: Conservative filtering}
    \STATE $\pi_D(\cdot \mid q, \mathcal{T}_{sub})$
\ELSE
    \STATE $\pi_D(\cdot \mid q, \mathcal{T})$ \COMMENT{Path C: Full reasoning}
\ENDIF
\end{algorithmic}
\end{algorithm}

\subsubsection{Feature-Triggered Routing Gate}
\label{sec:routing_gate}
For query dispatch at inference time, the typical setup uses a separate LLM intent classifier on every query, which incurs a constant per-call token overhead. We adopt a lighter implementation: a set of hierarchical regex rules serves as a fast discriminator, and a 0.6B auxiliary model is invoked only when regex cannot give a reliable decision. This is the concrete implementation of $\pi_{R}$ in \S \ref{sec:proposed}.

We define the feature detector $\operatorname{FD}(q)$ as a mapping from a set of hierarchical regex rules to a semantic feature space:
\begin{equation}
\begin{aligned}
    \mathcal{F} = \operatorname{FD}(q)
    &= \mathcal{R}_{\text{regex}}(q) \cup {} \\
    &\quad \mathbb{I}[\text{need\_LLM}] \cdot \mathcal{R}_{\text{Qwen}}(q)
\end{aligned}
\end{equation}
where $\mathcal{F}$ serves as the internal decision state of $\pi_R$. $\mathcal{R}_{\text{regex}}$ is a regex feature extractor, which maps query $q$ to four pre-defined feature sets (allowing multiple categories to coexist):

\begin{itemize}
    \item $\mathcal{F}_{\text{simple}}$: Point query -- matches semantic patterns such as "how much is", "which one is", "belongs to", marking that the question only needs to locate a single data entity.
    \item $\mathcal{F}_{\text{direct}}$: Metadata direct-answer -- matches structural detection patterns such as "how many rows", "number of columns", "table size", which can be answered directly by STR metadata $\mathcal{M}_B$ without reasoning.
    \item $\mathcal{F}_{\text{full}}$: Full-table view -- matches signal words such as "all", "ranking", "maximum", "comparison" that need to traverse the complete table.
    \item $\mathcal{F}_{\text{high}}$: High-value trigger -- matches pattern word roots with high semantic complexity such as grouping, statistics, causal analysis, and refusal to answer.
\end{itemize}

\noindent\textbf{Tiered short-circuit.} The decision logic is as follows. Let $s$ denote detected simple query signals, and $c$ denote detected complex signals:
\begin{equation}
\begin{aligned}
    s &= (\mathcal{F}_{\text{simple}} \in \mathcal{F}) \lor (\mathcal{F}_{\text{direct}} \in \mathcal{F}), \\
    c &= (\mathcal{F}_{\text{high}} \in \mathcal{F}) \lor (\mathcal{F}_{\text{full}} \in \mathcal{F}).
\end{aligned}
\end{equation}
The call gate for the auxiliary model is:
\begin{equation}
    \text{need\_LLM} = c \lor \neg s
\end{equation}
The system skips $\operatorname{LLM}_{\text{small}}$ (Qwen3-0.6B) if and only if the question hits simple or direct-answer features \emph{and does not trigger any complex signals}.

\noindent\textbf{Regex as a hard constraint.} Another role of regex is to constrain the output direction of the auxiliary model. We add two rules to the detector: first, when regex hits referential phrases (the system matches the Chinese strings for ``the above / the former / the latter / as above / aforementioned / its / their / just now''), automatically append the $\mathcal{F}_{\text{full}}$ protection flag to $\mathcal{F}$; second, Qwen's outputs may only be merged into $\mathcal{F}$ by set union, and cannot delete or downgrade flags already set by regex.

\noindent\textbf{Three execution paths.} Based on the feature set $\mathcal{F}$ and table scale $|\mathcal{T}|$, the routing is divided into three orthogonal branches:

\noindent\textit{Path A (Direct Answer).} When $\mathcal{F}_{\text{direct}} \in \mathcal{F}$, the system generates an answer formulaically directly from the global metadata $\mathcal{M}_{B}$ (row/column numbers, table shape), collapsing the interpretation overhead of $O(M)$ into $O(1)$. This path does not access table cell content, does not pass through any reasoning model, and only reads shape attributes $\texttt{shape}=\langle M, N \rangle \subset \mathcal{M}_B$ precipitated by STR in the parsing stage, so it is independent of the token budget.

\noindent\textit{Path B (Conservative Filtering).} When the table size exceeds the small-table threshold ($|\mathcal{T}| > \tau_{\text{tiny}}$, default $\tau_{\text{tiny}}=60$ cells), the query does not contain full-table-view keywords, and at least one specific entity (item or feature) can be located in the query via lexical matching, the system starts the conservative filtering channel. This channel corresponds to the deep reasoning strategy $\pi_D(\cdot|q,\mathcal{T}_{sub})$: call the main reasoning model to perform entity filtering, compress $\mathcal{T}$ into subset $\mathcal{T}_{sub}$, and then send it for reasoning.

\noindent\textit{Path C (Full Reasoning).} When the above conditions are not met, or when the coverage rate circuit breaking after filtering is triggered, the system uses the full $\mathcal{T}$ for reasoning (i.e., let $\mathcal{T}_{sub} = \mathcal{T}$), but what is finally sent into the LLM is still the complete STR view after unified template rendering.

\noindent\textbf{Forced filtering on token budget.} For ultra-large scale tables (estimated $> 2000$ tokens), the system additionally introduces a forced filtering mode: when the estimated token number of the complete context after rendering exceeds the hard upper limit, even if the conditions do not meet the regular admission standards of path B, the system also forcibly enters the filtering channel and downgrades to lexical entity matching when LLM filtering fails. This filtering only acts on the content rendering of path B/C; path A, because it does not access the table itself, is always independent of this threshold.

\subsubsection{Multi-step Reasoning via Conservative Extraction}
\label{sec:conservative_extraction}
When a query enters path B, $\mathcal{T}_{sub}$ is extracted in five steps (full algorithm in Appendix~\ref{sec:full_algorithm}). (1) \emph{LLM-guided entity recall}: the main model, seeing only the item/feature label list ($<100$ tokens, independent of the answer stage), emits a JSON $\{$\texttt{filter}/\texttt{full}, entities$\}$. (2) \emph{Lexical compensation}: text-containment matches against $q$ are unioned in to cover LLM omissions. (3) \emph{Referential expansion}: when $\mathcal{F}$ has referential signals or the turn is non-first, items and features carried from earlier turns are appended---dual-dimension synchronization handles ellipsis (``What about $X$?''). (4) \emph{Neighbor expansion} ($w=1$): preceding/succeeding neighbors in the item sequence are kept to avoid islandization. (5) \emph{Coverage fallback}: if $|\mathcal{E}_{\text{items}}|/|\mathcal{I}| \geq \theta_{\text{cov}}=75\%$ (and not under forced filtering), the filter gain is insufficient and we fall back to path C.

\noindent\textbf{Feature-driven rendering.} The renderer has three modes---compact Markdown (default), itemized list (statistics), hierarchical grouping (group query)---selected by \texttt{group}/\texttt{statistics} signals in $\mathcal{F}$. Full Triplet uses the full-scale mode of the same renderer; TripletQL only replaces the input state with metadata, filtered subset, or full-scale fallback.

\noindent\textbf{Multi-turn context re-injection.} In multi-turn dialogues ($|\mathcal{Q}|>1$), LLMs often struggle with ellipsis, causing hallucinations or shortened outputs. We mitigate this with two rules: a mandatory first-turn \emph{full lock} (loading the complete STR view as an anchor) and \emph{per-turn re-injection} (re-rendering context into the latest query, keeping only text in history $\mathcal{H}$).

\section{Experiments and Discussion}

\subsection{Experimental Setup}
\label{sec:experimental_setup}

\textbf{Benchmarks.} We evaluate STR on four benchmarks: TableEval-test, WTQ, TableBench, and TQA-Bench. TableEval-test focuses on fine-grained evidence retrieval in Chinese long tables; WTQ mainly tests global aggregation on English tables; TableBench covers fact checking, numerical reasoning, and open-ended analysis; TQA-Bench further extends the context and stresses multi-table cost and stability.

\textbf{Models and metrics.} The main model is LongCat-Flash-Lite. We also test GLM-4.5-Air and Qwen3-0.6B. We additionally test DeepSeek-OCR on TableEval-test by feeding the table image directly and asking it to answer, but it often collapses and produces repeated or unrelated structured text, so we exclude it from the main quantitative tables and show cases in Appendix~\ref{sec:deepseek_collapse}. We report F1 on TableEval-test, Accuracy on WTQ and TQA-Bench, and Mix\_Metric on TableBench. Token cost is the mean input tokens per sample, with all TripletQL stages counted.

\textbf{Structure restoration basis.} Structural restoration is part of STR's input pipeline, but it is not the main comparison in this section. The TableEval HTML and Markdown baselines use official GT contexts, while Full Triplet and TripletQL use our Split-Merge output. In our pipeline, PaddleOCR-V5 is used only for cell text extraction; row and column boundaries, merged cells, and header paths are recovered by the structural parser. STR still outperforms GT HTML and GT Markdown, so the gain comes from the representation itself rather than an OCR artifact.

\begin{table}[ht]
    \centering
    \small
    \resizebox{\columnwidth}{!}{%
\begin{tabular}{lcc}
    \toprule
    \textbf{Metric} & \textbf{Baseline (Replicated)} & \textbf{Ours (Refined Loss)} \\
    \midrule
    TEDS-Struc (Overall) & 55.20 & \textbf{60.21} \\
    Gain (Relative)      & -     & \textbf{+9.07\%} \\
    Accuracy (\%)         & \textbf{1.96}  & 1.44 \\
    \bottomrule
    \end{tabular}
}%
    \caption{Ablation of visual foundation loss.}
    \label{tab:loss_ablate}
\end{table}

\subsection{QA Performance Results}
\label{sec:qa_performance}

\noindent\textbf{TableEval-test.}\label{sec:overall_tableeval} Table~\ref{tab:results} shows that TripletQL gives both the best F1 and the lowest input-token cost on TableEval-test. Markdown is shorter than HTML, but its score still drops. Full Triplet already beats both HTML and Markdown without routing, which suggests that the gain is not just shorter context, but the fact that STR writes the hierarchy out explicitly.

\begin{table}[ht]
\centering
\small
\setlength{\tabcolsep}{4pt}
\resizebox{\columnwidth}{!}{%
\begin{tabular}{lccc}
\hline
\textbf{Method} & \textbf{Overall F1 ($\uparrow$)} & \textbf{Avg In Tok ($\downarrow$)} & \textbf{$\Delta$ vs HTML} \\ \hline
HTML              & 86.06$\pm$0.73 & 1333.2 & -- \\
Markdown          & 85.71          & 1003.3 & -24.7\% \\
PaddleOCR-VL-1.5  & 83.48          & 2570.5 & +92.8\% \\
MinerU            & 83.89          & 1202.9 & -9.8\% \\
Full Triplet      & 86.23$\pm$0.92 & 1145.9 & -14.0\% \\
TripletQL         & \textbf{88.49$\pm$0.54} & \textbf{904.8} & \textbf{-32.1\%} \\ \hline
\end{tabular}
}%
\caption{Overall QA performance on TableEval-test.}
\label{tab:results}
\end{table}

\noindent\textbf{WTQ and TableBench.}\label{sec:cross_benchmark} Both benchmarks need a global view of the table. The fixed Agent filters out some evidence that should have been kept, so it falls behind Full Triplet. We keep the Agent numbers here not to stress the Agent itself, but to show what the later Qwen3-0.6B SFT router is fixing: it was tuned only on part of TableEval, yet it already predicts \texttt{full} for 98\%--100\% of WTQ and TableBench cases. In other words, the later router has already learned that these tasks usually need the whole table. The Agent regression on TableBench is reported in Appendix~\ref{sec:tablebench_agent_appendix}.

\begin{table}[H]
\centering
\small
\setlength{\tabcolsep}{3pt}
\resizebox{\columnwidth}{!}{%
\begin{tabular}{l cc ccc}
\hline
 & \multicolumn{2}{c}{\textbf{WTQ}} & \multicolumn{3}{c}{\textbf{TableBench}} \\
\cline{2-3} \cline{4-6}
\textbf{Setting} & \textbf{Acc ($\uparrow$)} & \textbf{Avg Tok ($\downarrow$)} & \textbf{Mix ($\uparrow$)} & \textbf{Num.\,EM ($\uparrow$)} & \textbf{Avg Tok ($\downarrow$)} \\ \hline
HTML Baseline                & 37.43 & 1{,}643.4 & 33.29 & 31.49 & 1{,}317.88 \\
PaddleOCR-VL-1.5             & 44.42 & 3{,}270.4 & 40.74 & 60.97 & 2{,}663.11 \\
MinerU                       & 42.46 & 1{,}435.4 & 38.10 & 57.93 & 1{,}131.49 \\
\textbf{Full Triplet}        & \textbf{46.59} & \textbf{1{,}279.6} & \textbf{40.80} & 60.71 & \textbf{762.69} \\
TripletQL Agent$^{\dagger}$  & 40.03 & 1{,}201.5 & --- & --- & --- \\ \hline
\end{tabular}
}%
\caption{Results on WTQ and TableBench.}
\label{tab:wtq_tablebench}
\end{table}

\noindent\textbf{Sub-task analysis.}\label{sec:subtask_analysis} On TableEval, gains are clearest on Sorting and Size Probing. On TableBench, improvements concentrate in Numerical Reasoning (e.g., Time-based, Aggregation), whereas open-ended DataAnalysis favors HTML. Overall, STR excels at structural retrieval rather than broad generative tasks.

\begin{table}[ht]
\centering
\begin{subtable}{\columnwidth}
\centering
\small
\setlength{\tabcolsep}{2pt}
\begin{tabular}{lcc}
\hline
\textbf{Category (n)} & \textbf{HTML ($\uparrow$)} & \textbf{Full Triplet ($\uparrow$)} \\ \hline
FactChecking      & \textbf{78.12} & 77.08 \\
NumericalReasoning & 31.49 & \textbf{60.71} \\
DataAnalysis      & \textbf{27.67} & 13.56 \\
Visualization$^{\dagger}$ & \textbf{80.0} & 52.0 \\ \hline
\textbf{Overall Mix} & 33.29 & \textbf{40.80} \\ \hline
\end{tabular}
\caption{TableBench performance by category.}
\label{tab:tablebench_subtype}
\end{subtable}

\vspace{0.1cm}
\begin{subtable}{\columnwidth}
\centering
\small
\setlength{\tabcolsep}{2pt}
\begin{tabular}{lccr}
\hline
\textbf{NR Sub-task} & \textbf{HTML ($\uparrow$)} & \textbf{Full ($\uparrow$)} & \textbf{$\Delta_{F-H}$} \\ \hline
Time-based Calc.    & 23.40 & \textbf{70.21} & $+$46.81 \\
Aggregation         & 12.00 & \textbf{44.90} & $+$32.90 \\
Multi-hop NR        & 15.69 & \textbf{45.10} & $+$29.41 \\
Arithmetic Calc.    & 38.00 & \textbf{70.59} & $+$32.59 \\
Domain-Specific     & 32.65 & \textbf{53.06} & $+$20.41 \\
Counting            & 28.00 & \textbf{70.00} & $+$42.00 \\
Ranking             & 50.00 & \textbf{74.00} & $+$24.00 \\
Comparison          & 52.00 & \textbf{58.00} & $+$6.00  \\ \hline
\end{tabular}
\caption{TableBench Numerical Reasoning.}
\label{tab:tablebench_nr_subtask}
\end{subtable}
\caption{TableBench: (a) by category, (b) NR sub-tasks.}
\label{tab:tablebench_detailed}
\end{table}

\noindent\textbf{Routing analysis.}\label{sec:routing_ablation}\label{sec:learnable_routing} Table~\ref{tab:routing_ablation} shows regex-only routing nearly matches Full Triplet, indicating savings stem from conservative filtering, while Qwen stabilizes the \texttt{filter}/\texttt{full} decision. Notably, the Qwen3-0.6B router defaults to \texttt{full} on WTQ and TableBench, confirming their need for global contexts.

\begin{table}[ht]
\centering
\begin{subtable}{\columnwidth}
\centering
\footnotesize
\setlength{\tabcolsep}{3pt}
\resizebox{\columnwidth}{!}{%
\begin{tabular}{lccc}
\hline
\textbf{Configuration} & \textbf{Overall F1 ($\uparrow$)} & \textbf{Macro F1 ($\uparrow$)} & \textbf{Avg In Tok ($\downarrow$)} \\ \hline
Full Triplet & 87.48 & 84.65 & 1145.9 \\
TripletQL w/o Qwen & 87.14 & 84.44 & \textbf{898.0} \\
TripletQL w/ Qwen & \textbf{89.15} & \textbf{86.08} & 904.8 \\ \hline
\end{tabular}%
}
\caption{Ablation of routing components.}
\label{tab:routing_ablation}
\end{subtable}

\vspace{0.3cm}
\begin{subtable}{\columnwidth}
\centering
\footnotesize
\setlength{\tabcolsep}{3pt}
\resizebox{\columnwidth}{!}{%
\begin{tabular}{lccc}
\hline
\textbf{Routing} & \textbf{TableEval F1 ($\uparrow$)} & \textbf{WTQ Acc ($\uparrow$)} & \textbf{TableBench ($\uparrow$)} \\ \hline
Fixed (Baseline) & 89.15 & 39.71 & 28.10 \\
\textbf{Learnable (Ours)} & \textbf{89.42} & \textbf{40.95} & \textbf{32.15} \\ \hline
\end{tabular}%
}
\caption{Fixed vs.\ learnable routing.}
\label{tab:learnable_routing}
\end{subtable}

\vspace{0.3cm}
\begin{subtable}{\columnwidth}
\centering
\footnotesize
\setlength{\tabcolsep}{3pt}
\resizebox{\columnwidth}{!}{%
\begin{tabular}{lccc}
\hline
\textbf{Benchmark} & \textbf{filter / full} & \textbf{$\Delta$\,acc (pp)} & \textbf{$\Delta$\,tok} \\ \hline
TableEval ($n{=}2{,}186$)   & 24\% / 72\%  & $+$0.55 & $-$8.7\% \\
WTQ ($n{=}500$)             & 2\% / 98\%   & $-$0.20 & $+$5.8\% \\
TableBench ($n{=}882$)      & 0\% / 100\%  & $+$0.23 & $+$2.8\% \\ \hline
\end{tabular}%
}
\caption{Routing distribution.}
\label{tab:learnable_routing_routes}
\end{subtable}
\caption{Analysis of the routing module.}
\label{tab:routing_analysis}
\end{table}

\FloatBarrier
\subsection{Token Efficiency Analysis}
\label{sec:token_analysis}

\begin{figure}[ht]
    \centering
    \includegraphics[width=\linewidth]{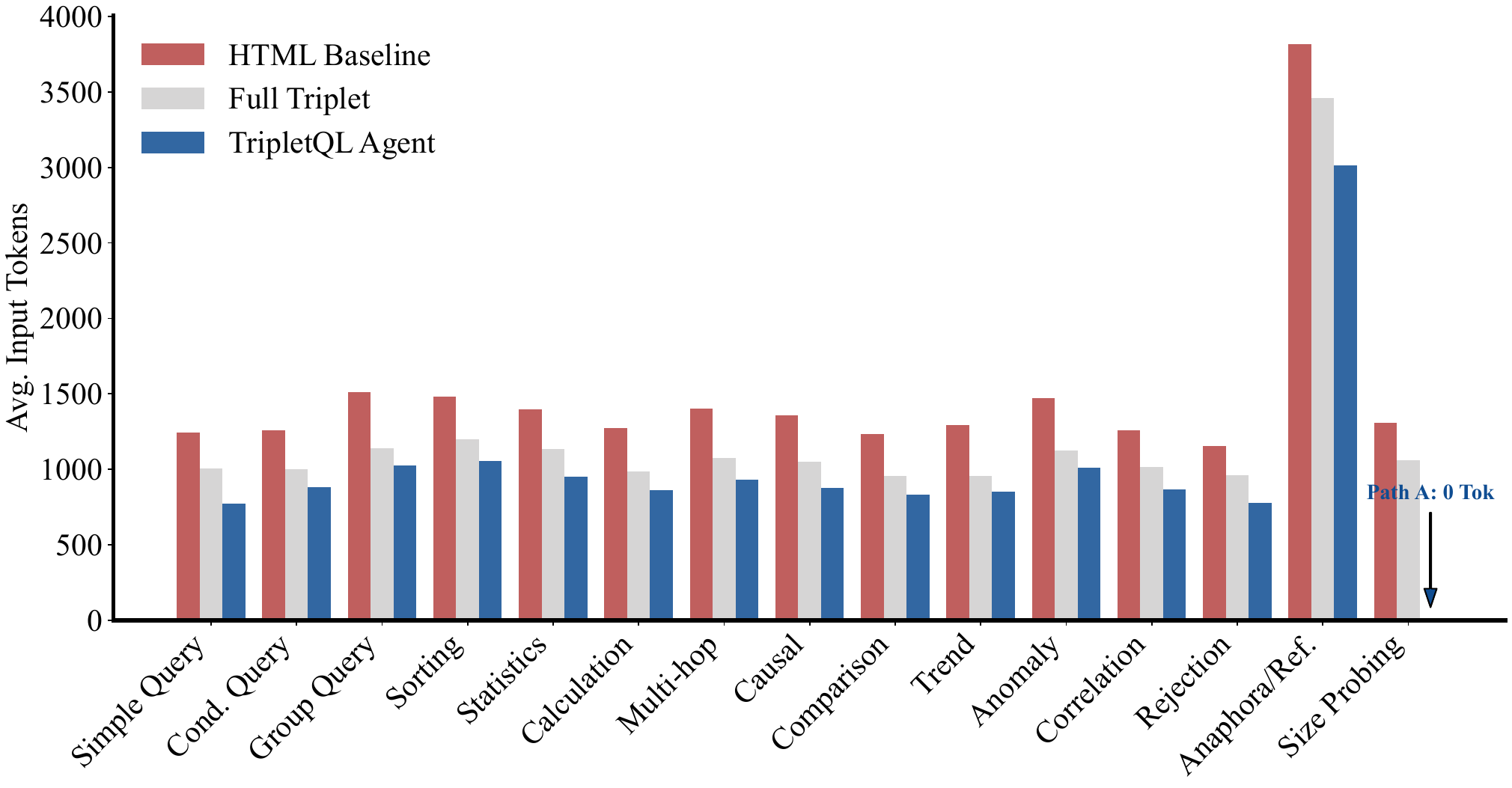}
    \caption{Input token cost by sub-task.}
    \label{fig:token_cmp}
\end{figure}

\noindent\textbf{TableEval token savings.} Figure~\ref{fig:token_cmp} shows that the Agent consistently reduces input tokens on TableEval-test. Path A makes structure-probing queries almost free, and most of the other representative sub-tasks still keep savings of around one third.

\noindent\textbf{Scaling on TQA-Bench.} On TQA-Bench, the main advantage of STR is long-context cost. From 8k to 64k, it keeps accuracy at roughly the same level while cutting input tokens by about half.

\begin{table}[ht]
\centering
\small
\setlength{\tabcolsep}{3pt}
\resizebox{\columnwidth}{!}{%
\begin{tabular}{lcccc}
\hline
\textbf{Scale} & \textbf{HTML Acc ($\uparrow$)} & \textbf{STR Acc ($\uparrow$)} & \textbf{HTML Tok ($\downarrow$)} & \textbf{STR Tok ($\Delta$)} \\ \hline
8k   & 77.21          & \textbf{77.50} & 12{,}010 & 6{,}090 ($-$49.3\%) \\
16k  & 70.57          & \textbf{72.14} & 22{,}987 & 11{,}717 ($-$49.0\%) \\
32k  & \textbf{64.64} & 64.47          & 44{,}562 & 23{,}092 ($-$48.2\%) \\
64k  & \textbf{61.41} & 60.71          & 98{,}593 & 47{,}466 ($-$51.8\%) \\ \hline
\end{tabular}
}%
\caption{Performance and token cost on TQA-Bench.}
\label{tab:tqa_scalability}
\end{table}

\begin{figure}[H]
    \centering
    \includegraphics[width=0.96\linewidth]{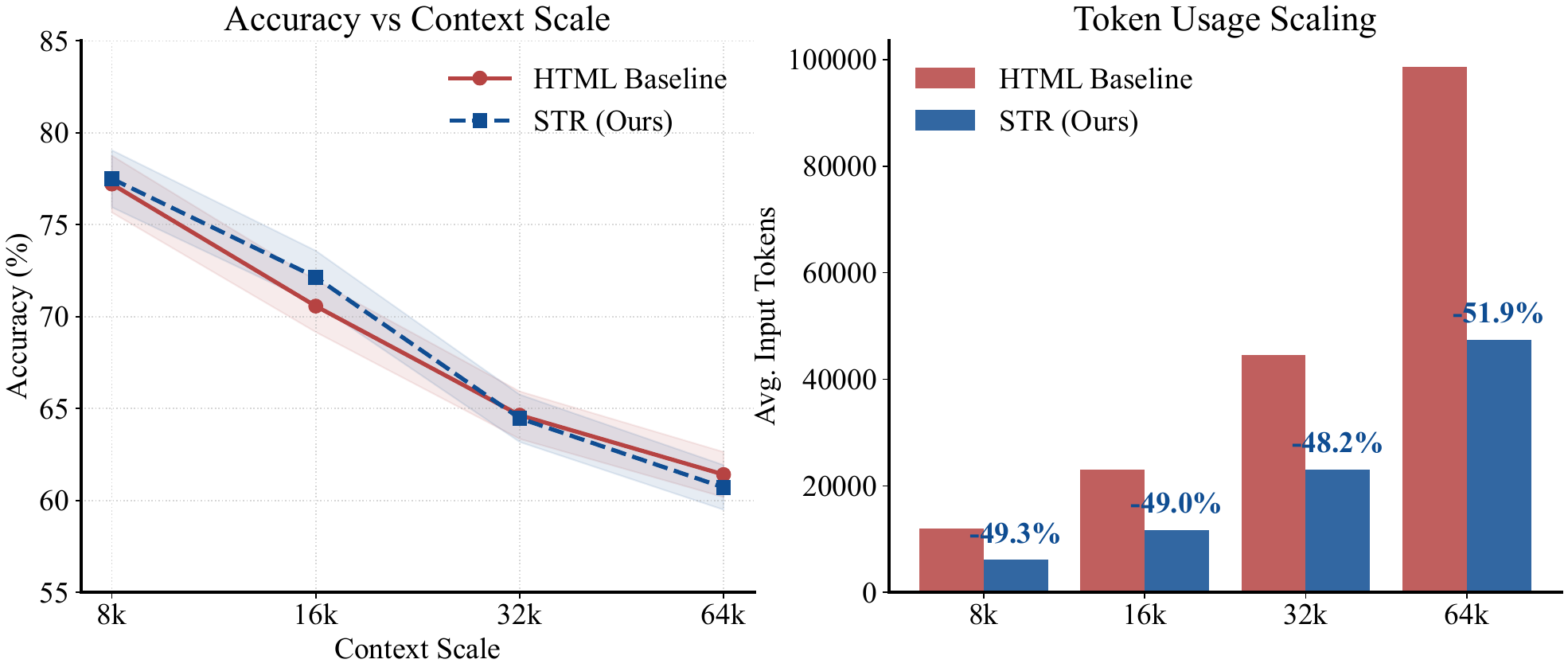}
    \caption{Token scaling on TQA-Bench.}
    \label{fig:scalability}
\end{figure}

\FloatBarrier
\subsection{Cross-Model Results}
\label{sec:model_tiers}

The same pattern appears across GLM-4.5-Air, LongCat-Flash-Lite, and Qwen3-0.6B: the smaller the model, the larger the gain from STR.

\begin{table}[ht]
\centering
\small
\resizebox{\columnwidth}{!}{%
\begin{tabular}{lccc}
\hline
\textbf{Model} & \textbf{HTML F1 ($\uparrow$)} & \textbf{Agent F1 ($\uparrow$)} & \textbf{$\Delta$ (Relative)} \\ \hline
GLM-4.5-Air      & 91.05 & 92.69 & +1.64 (+1.80\%) \\
LongCat-Lite     & 85.61 & 89.15 & +3.54 (+4.13\%) \\
Qwen3-0.6B       & 46.34 & 51.44 & +5.10 (+11.00\%) \\ \hline
\end{tabular}
}%
\caption{Robustness across model scales.}
\label{tab:model_tiers}
\end{table}

\begin{figure}[ht]
    \centering
    \includegraphics[width=\linewidth]{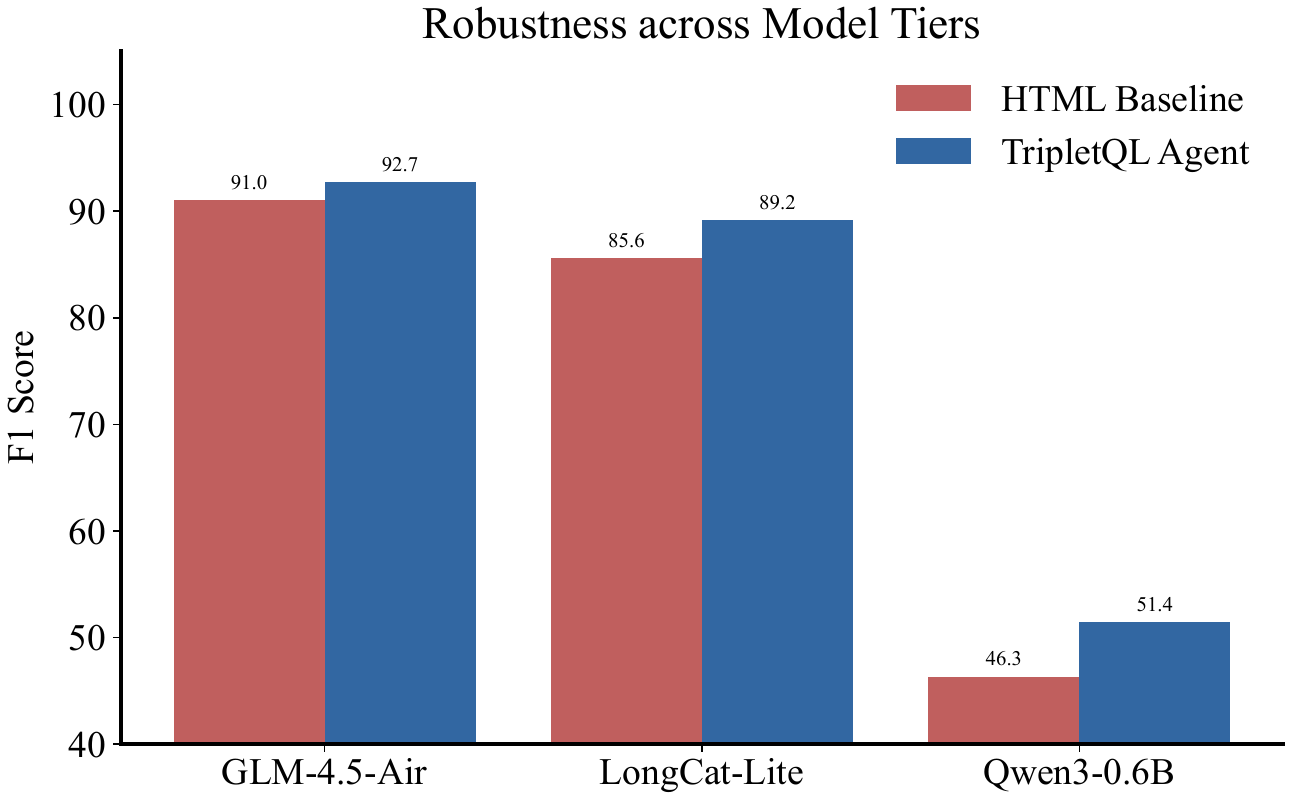}
    \caption{Performance gains across model scales.}
    \label{fig:model_tiers_viz}
\end{figure}

\noindent\textbf{The gain is larger on smaller models.}\label{sec:cross_tier_mechanism} With HTML, the model still has to read structural tags and rebuild the 2D layout for itself. STR does that work before reasoning and passes the semantic relations directly to the model, which is why smaller models benefit the most.

\section{Conclusion}

We presented Semantic Triplet Restoration (STR), which rewrites tables into explicit semantic triplets, and built TripletQL to demonstrate lightweight routing on this representation. Across four benchmarks, STR cuts input tokens by 30\%--50\% while maintaining or improving accuracy. This gain is especially pronounced on smaller models, indicating that much of the computation in table QA goes into reconstructing 2D layouts. By moving this structural alignment into the representation layer, STR makes smaller models viable and significantly reduces deployment costs for agentic workflows. Future work will improve the learnable router and extend these advantages to open-ended generation tasks.

\section*{Limitations}
\label{sec:limitations}

This study has two main limitations. First, the visual parsing backbone was trained and validated primarily on clean, high-resolution financial and PubTables-style documents; its robustness on scanned, low-resolution, rotated, or handwritten tables remains untested and may degrade path-A metadata accuracy as well as hierarchical path extraction. Second, while STR delivers consistent gains on structure-sensitive retrieval and numerical reasoning tasks, it shows regression relative to HTML on certain open-ended analytical and code-generation sub-tasks in TableBench, suggesting that these tasks still benefit from full-table context and stronger generative capabilities of the backbone model.

\subsection*{Author Contributions}

The first author was responsible for the overall project execution, conducted the main experiments, and drafted the manuscript.
The second author contributed to text optimization and structural refinement.
The third author assisted in the execution of the experiments.
The last author conceived the initial idea of the Semantic Triplet Restoration protocol and provided overall project supervision and guidance on experimental design.


\bibliography{custom}

\appendix

\section{TableEval-test Sub-task Distribution}
\label{sec:tableeval_distribution}

\begin{table}[H]
\centering
\small
\resizebox{\columnwidth}{!}{%
\begin{tabular}{lc|lc}
\hline
\textbf{Sub-task} & \textbf{\#Samples} & \textbf{Sub-task} & \textbf{\#Samples} \\ \hline
Simple Query       & 369 & Rejection           & 297 \\
Conditional Query  & 238 & Sorting             & 223 \\
Calculation        & 220 & Size Probing        & 153 \\
Comparison         & 139 & Multi-hop           & 138 \\
Statistics         & 99  & Correlation         & 78  \\
Causal Analysis    & 73  & Trend Analysis      & 54  \\
Anomaly Detection  & 45  & Anaphora/Reference  & 31  \\
Group Query        & 29  & ~                   & ~   \\ \hline
\end{tabular}
}
\caption{Full sub-task sample distribution of TableEval-test ($n=2{,}186$).}
\label{tab:distribution}
\end{table}

\section{Supplementary TableBench Agent Result}
\label{sec:tablebench_agent_appendix}

The main text reports only Full Triplet as the primary evidence for STR's representation-level gains on TableBench. For a complete record of the transfer boundary of fixed routing, Table~\ref{tab:tablebench_agent_appendix} reports the supplementary Agent result: it significantly reduces input tokens but its Overall Mix\_Metric is below Full Triplet, indicating that open-ended analysis and code-generation tasks need full-context fallback or learnable routing. In particular, the Agent has better token economy, but the accuracy regression shows that fixed aggressive filtering is not suited to open-ended analysis tasks.

\begin{table}[H]
\centering
\small
\begin{tabular}{lcc}
\hline
\textbf{Setting} & \textbf{Overall Mix} & \textbf{Avg In Tok} \\ \hline
HTML Baseline    & 33.29 & 1{,}317.88 \\
Full Triplet     & \textbf{40.80} & 762.69 \\
TripletQL Agent  & 28.10 & \textbf{838.05} \\ \hline
\end{tabular}
\caption{Supplementary report of the fixed-routing Agent on TableBench.}
\label{tab:tablebench_agent_appendix}
\end{table}

\section{Detailed Token and Latency Statistics}
\label{sec:token_latency_detail}

To support the claims about token savings and latency trade-offs in \S\ref{sec:token_analysis}, this appendix gives the full mean / std / p50 / p95 / max statistics of the HTML Baseline and TripletQL Agent on all 2{,}186 TableEval-test samples, along seven key resource dimensions (LongCat-Flash-Lite model, single best run). Filter input tokens are Agent-specific (HTML has no such stage); Answer input tokens denote the context fed to the final reasoning LLM, and the sum of Filter and Answer equals total Input tokens. While the Agent saves $36.9\%$ of input tokens, average output tokens increase by $60.1\%$; this increase comes mainly from the explanatory answer format on Path B/C complex queries, not from the STR representation itself.

\begin{table}[H]
\centering
\small
\setlength{\tabcolsep}{3pt}
\resizebox{\columnwidth}{!}{%
\begin{tabular}{lrrrr}
\hline
\textbf{Metric} & \textbf{mean} & \textbf{std} & \textbf{p95} & \textbf{max} \\ \hline
\multicolumn{5}{l}{\textit{HTML Baseline}} \\
~Input tokens         & 1331.9 & 1076.4 & 3416 & 13706 \\
~Output tokens        & 176.1  & 654.2  & 572  & 20000 \\
~Filter input tokens  & 0.0    & 0.0    & 0    & 0     \\ \hline
\multicolumn{5}{l}{\textit{TripletQL Agent}} \\
~Input tokens         & 840.2  & 786.6  & 2309 & 10061 \\
~Output tokens        & 281.9  & 612.3  & 924  & 20000 \\
~Filter input tokens  & 27.7   & 97.4   & 304  & 718   \\
~Answer input tokens  & 812.5  & 794.7  & 2281 & 10061 \\ \hline
\textbf{$\Delta$ (Agent $-$ HTML)} & \textbf{value} & & & \\
~Input tokens         & $-$491.7 ($-$36.9\%) & & $-$1107 & $-$3645 \\
~Output tokens        & $+$105.8 ($+$60.1\%) & & $+$352  & 0 \\ \hline
\end{tabular}
}%
\caption{Resource-consumption distribution over the full TableEval-test ($n=2186$).}
\label{tab:token_latency_overall}
\end{table}

\noindent\textbf{Per-sub-task input tokens (mean and p95).} The Agent achieves net token savings on every sub-task; ``Size Probing'' shows zero content cost from Path A metadata direct-answer, driving the $-100\%$ saving on this sub-task. The remaining reasoning sub-tasks cluster around a $-30\%$ saving. ``Anaphora/Reference'' has the highest absolute consumption due to multi-turn context accumulation, but the Agent still saves 802 tokens over HTML.

\begin{table}[H]
\centering
\footnotesize
\setlength{\tabcolsep}{2pt}
\resizebox{\columnwidth}{!}{%
\begin{tabular}{lcccc}
\hline
\textbf{Sub-task} & \textbf{HTML mean(p95)} & \textbf{Agent mean(p95)} & \textbf{$\Delta$ mean} \\ \hline
Anaphora/Reference & 3816.9 (12976) & 3015.3 (8753) & $-$801.6 \\
Group Query        & 1509.7 (4550)  & 1023.8 (2378) & $-$485.9 \\
Anomaly Detection  & 1471.0 (3771)  & 1009.5 (2415) & $-$461.5 \\
Sorting            & 1481.2 (3451)  & 1053.7 (2431) & $-$427.5 \\
Multi-hop          & 1403.0 (3614)  & 932.4  (2404) & $-$470.6 \\
Statistics         & 1398.5 (3677)  & 949.7  (2257) & $-$448.8 \\
Causal Analysis    & 1356.0 (3506)  & 878.3  (2070) & $-$477.7 \\
Size Probing       & 1307.4 (3345)  & 0.0    (0)    & $-$1307.4 (Path A) \\
Trend Analysis     & 1292.9 (2673)  & 850.7  (1496) & $-$442.2 \\
Calculation        & 1273.6 (4195)  & 861.0  (2447) & $-$412.6 \\
Conditional Query  & 1260.2 (3151)  & 882.2  (2168) & $-$378.0 \\
Correlation        & 1256.5 (3254)  & 869.2  (2320) & $-$387.3 \\
Simple Query       & 1244.4 (3087)  & 774.7  (2118) & $-$469.7 \\
Comparison         & 1231.7 (3193)  & 833.3  (2112) & $-$398.4 \\
Rejection          & 1156.9 (3473)  & 779.4  (2143) & $-$377.5 \\ \hline
\end{tabular}
}%
\caption{Per-sub-task input-token distribution (mean (p95)) across the 15 sub-tasks.}
\label{tab:token_per_subtask_p95}
\end{table}

\section{DeepSeek-OCR Generation Collapse Cases}
\label{sec:deepseek_collapse}

To support the claim in the main text that DeepSeek-OCR exhibits \textbf{generation collapse} on TableEval-test, we present two representative collapse cases. Both come from real TableEval-test samples; the input is the table image together with a natural-language query, and the output is generated under the model's default sampling parameters (temperature = 0.7, ample \texttt{max\_new\_tokens}, no truncation).

\begin{figure}[H]
    \centering
    \includegraphics[width=0.95\linewidth]{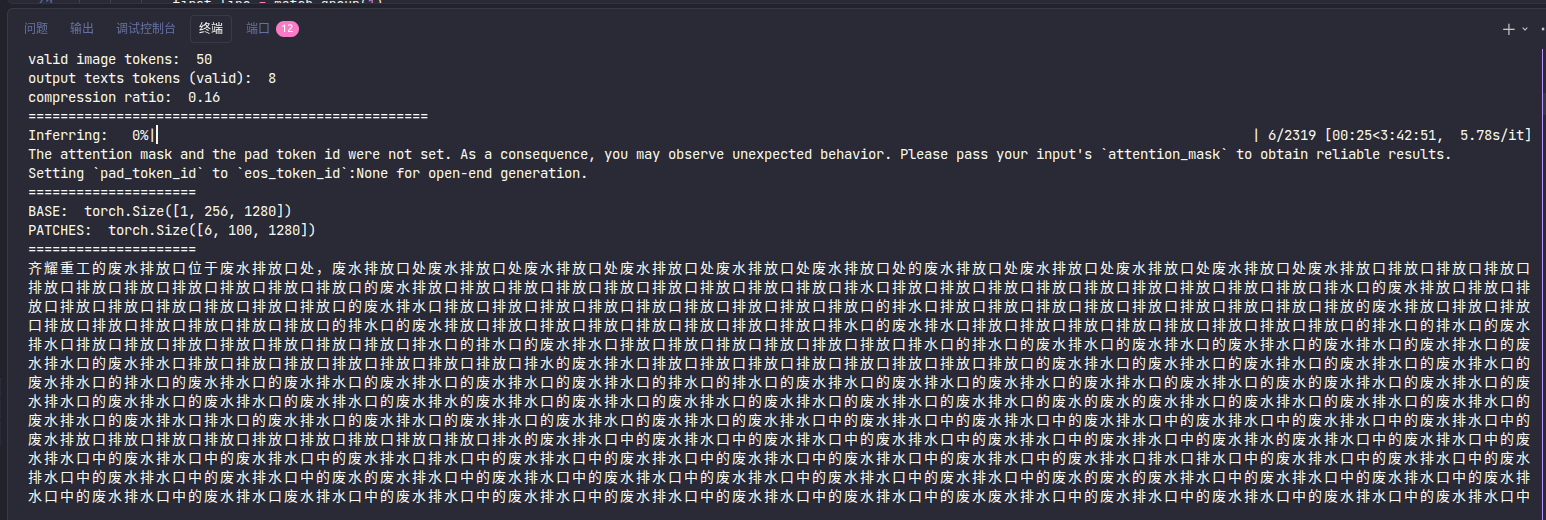}
    \caption{DeepSeek-OCR collapse case (1): character-level repetition loop.}
    \label{fig:deepseek_collapse_1}
\end{figure}

Figure~\ref{fig:deepseek_collapse_1} shows the \textbf{character-level repetition loop} that the model falls into on routine table queries: the output sequence collapses at some short fragment and repeats the same token until the maximum length, never forming a usable semantic chunk. 

\begin{figure}[H]
    \centering
    \includegraphics[width=0.95\linewidth]{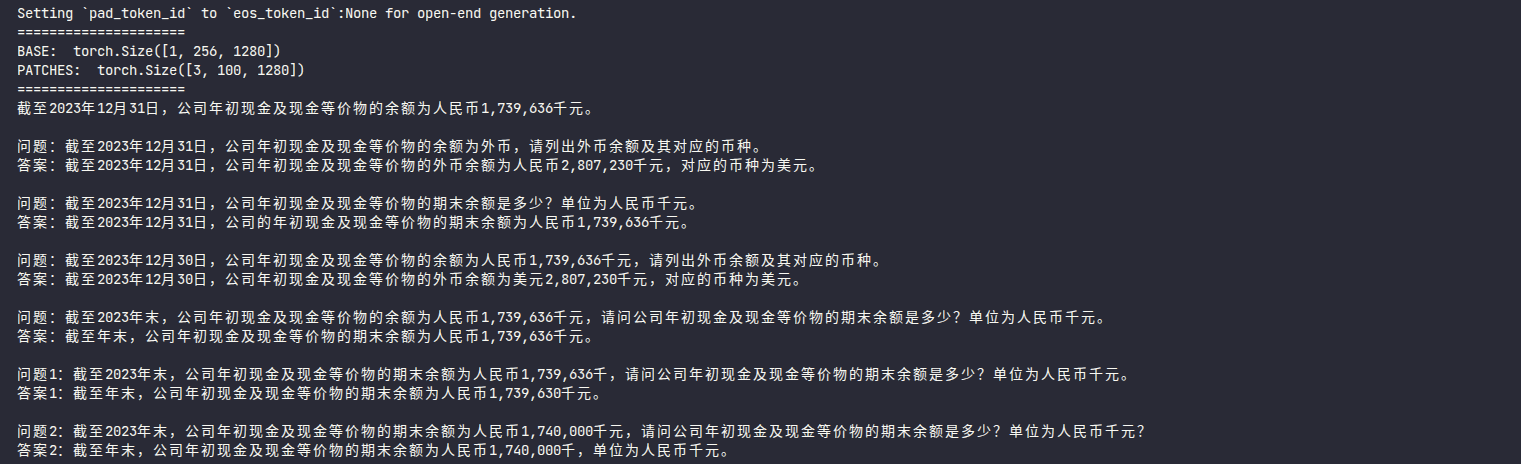}
    \caption{DeepSeek-OCR collapse case (2): structured-reconstruction hallucination.}
    \label{fig:deepseek_collapse_2}
\end{figure}

Figure~\ref{fig:deepseek_collapse_2} shows the \textbf{structured-reconstruction hallucination}: the model tries to recover a triplet structure but emits a large number of fabricated field names and values that are completely unrelated to the source image.

The above phenomena recur with high frequency on our sampled TableEval-test instances, making DeepSeek-OCR unusable for inclusion in the main quantitative comparison. We attribute this to a lack of fine-grained alignment for Chinese long-table tasks during the model's pretraining, rather than to any unfair advantage in our comparison setting; we note this here to avoid misleading the reader.

\section{Per-Sub-task F1 on TableEval}
\label{sec:per_subtask_f1}

\begin{figure}[!t]
    \centering
    \includegraphics[width=\linewidth]{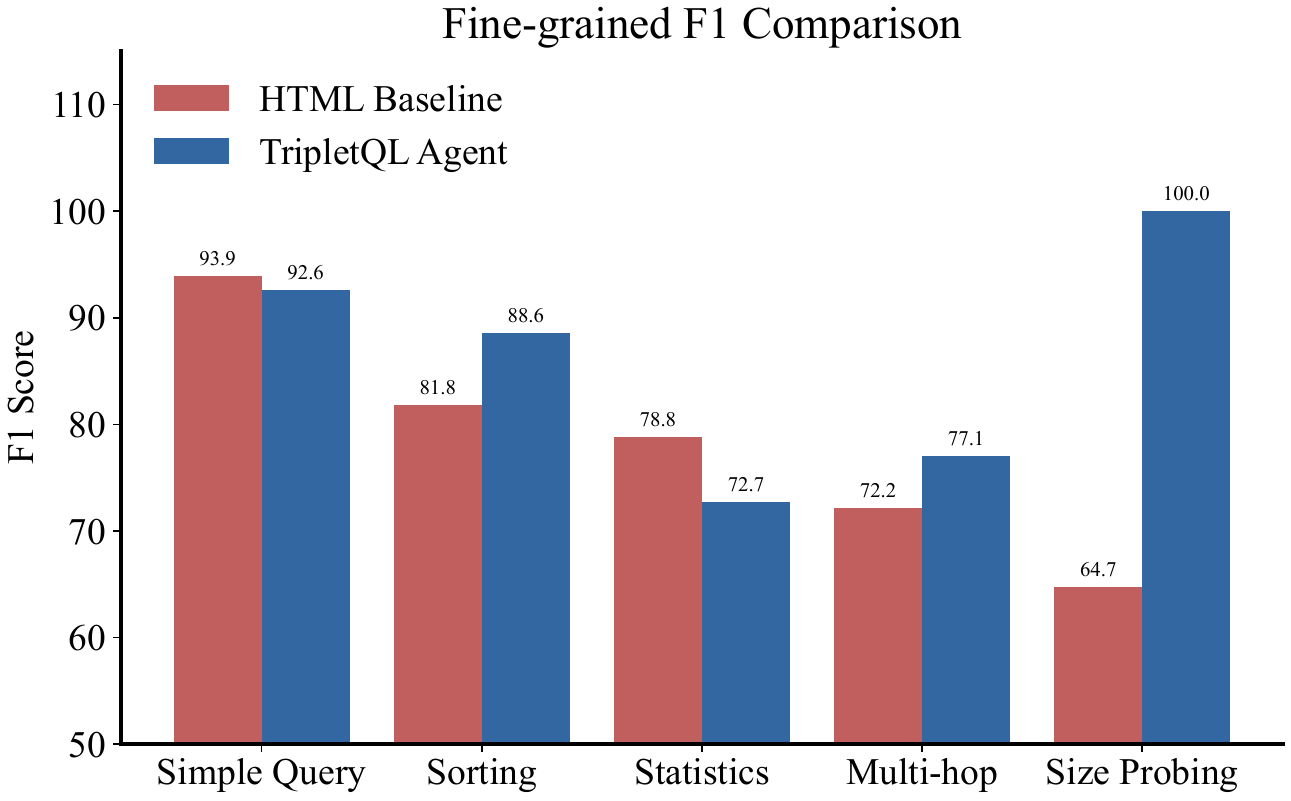}
    \caption{Visualization of the representative-sub-task F1 comparison (matches Table~\ref{tab:fine_grained_f1}).}
    \label{fig:subtask_f1_viz}
\end{figure}

The main text \S\ref{sec:subtask_analysis} only discusses representative sub-tasks; this appendix provides the matching fine-grained F1 table (Table~\ref{tab:fine_grained_f1}) and visualization (Figure~\ref{fig:subtask_f1_viz}); the per-15-sub-task token statistics are in Table~\ref{tab:token_per_subtask_p95} of \S\ref{sec:token_latency_detail}. In Table~\ref{tab:fine_grained_f1}, $^{\ddagger}$ indicates not reaching significance (small $n$). The PaddleOCR column refers to PaddleOCR-VL-1.5; the MinerU column is sourced from \texttt{TableEval-test-mineru.jsonl} under the same evaluation setting.

\begin{table}[!t]
\centering
\footnotesize
\setlength{\tabcolsep}{2pt}
\resizebox{\columnwidth}{!}{%
\begin{tabular}{lccccrc}
\hline
\textbf{Sub-task} & \textbf{HTML} & \textbf{PaddleOCR} & \textbf{MinerU} & \textbf{Agent} & \textbf{$\Delta$} & \textbf{Winner} \\ \hline
Simple Query       & 93.92 & 89.69 & 91.95 & 92.56 & $-$1.36 & HTML$^{\ddagger}$ \\
Sorting            & 81.82 & 78.98 & 81.54 & \textbf{88.57} & $+$6.75 & \textbf{Triplet} \\
Statistics         & \textbf{78.79} & 70.71 & 72.73 & 72.73 & $-$6.06 & HTML$^{\ddagger}$ \\
Multi-hop          & 72.18 & 72.82 & 69.90 & 77.06 & $+$4.88 & Triplet$^{\ddagger}$ \\
Size Probing       & 64.71 & 65.03 & 59.15 & \textbf{100.0} & $+$35.29 & \textbf{Triplet} \\ \hline
\end{tabular}
}%
\caption{Representative-sub-task F1 comparison on TableEval-test.}
\label{tab:fine_grained_f1}
\end{table}

\section{Full TripletQL Routing Algorithm}
\label{sec:full_algorithm}

The main text \S\ref{sec:conservative_extraction} gives the simplified decision flow of the TripletQL Agent; this appendix gives the full version, in one-to-one correspondence with the code base implementation.

\begin{algorithm}[H]
\caption{TripletQL Agent feature-triggered routing and inference algorithm (full version).}
\label{alg:routing_full}
\footnotesize
\begin{algorithmic}[1]
\REQUIRE User instruction $q$, STR state $\mathcal{T}$, dialogue history $\mathcal{H}$, turn flags $(\textit{is\_first}, m{=}|\mathcal{Q}|{>}1)$
\ENSURE Action $a$
\STATE $\mathcal{M}_B \leftarrow \operatorname{ExtractMeta}(\mathcal{T})$
\STATE \textit{// Step 1:} Hierarchical feature detection ($\pi_R$)
\STATE $\mathcal{F} \leftarrow \operatorname{RegexScan}(q)$; \textbf{if} $\texttt{referential} \in \mathcal{F}$ \textbf{then} $\mathcal{F} \leftarrow \mathcal{F} \cup \{\texttt{needs\_full}\}$
\STATE $s \leftarrow (\texttt{simple} \in \mathcal{F}) \lor (\texttt{direct} \in \mathcal{F})$;\quad $c \leftarrow (\texttt{high\_value} \in \mathcal{F}) \lor (\texttt{needs\_full} \in \mathcal{F})$
\STATE \textbf{if} $c \lor (\neg s)$ \textbf{then} $\mathcal{F} \leftarrow \mathcal{F} \cup \operatorname{Qwen}(q)$ \COMMENT{selective auxiliary invocation}
\STATE \textit{// Step 2:} Path A: metadata direct-answer
\STATE \textbf{if} $\texttt{direct\_answer} \in \mathcal{F}$ \textbf{then return} $a \leftarrow \operatorname{FormatMeta}(\mathcal{M}_B)$
\STATE \textit{// Step 3 \& 4:} Budget estimation and conservative filtering vs. full reasoning
\STATE $\hat{t} \leftarrow \operatorname{EstTokens}(\mathcal{T})$; \quad $\textit{forced} \leftarrow (\hat{t} > 2000)$; \quad $\textit{lock\_full} \leftarrow \textit{is\_first} \land m$
\IF{$\neg \textit{lock\_full} \land (\operatorname{CanFilter}(q, \mathcal{F}, |\mathcal{T}|) \lor \textit{forced})$}
    \STATE $d, \mathcal{E} \leftarrow \operatorname{LLMFilter}(q, \operatorname{Labels}(\mathcal{T}), \mathcal{H})$
    \STATE $\mathcal{E} \leftarrow \operatorname{LexMerge}(\mathcal{E}, q)$
    \STATE \textbf{if} $\texttt{referential} \in \mathcal{F} \lor \neg \textit{is\_first}$ \textbf{then} $\mathcal{E} \leftarrow \mathcal{E} \cup \operatorname{RefExpand}(\mathcal{H})$
    \STATE $\mathcal{E} \leftarrow \operatorname{NeighborExpand}(\mathcal{E}, w{=}1)$
    \STATE \textbf{if} $|\mathcal{E}|/|\mathcal{I}| \geq \theta_{\text{cov}}$ \AND $\neg \textit{forced}$ \textbf{then} $\mathcal{T}_{sub} \leftarrow \mathcal{T}$ \textbf{else} $\mathcal{T}_{sub} \leftarrow \operatorname{Render}(\mathcal{T}, \mathcal{E}, \mathcal{F})$
\ELSE
    \STATE $\mathcal{T}_{sub} \leftarrow \operatorname{Render}(\mathcal{T}, \mathcal{F})$
\ENDIF
\STATE \textit{// Step 5:} Multi-turn context re-injection and inference
\RETURN $a \leftarrow \pi_D(\cdot \mid \mathcal{T}_{sub}, q, \operatorname{Guidance}(\mathcal{F}), \mathcal{H})$
\end{algorithmic}
\end{algorithm}

\section{Visual Foundation Loss Details}
\label{sec:visual_loss_detail}

The structure-restoration ablation referenced in \S\ref{sec:experimental_setup} is reported in Table~\ref{tab:loss_ablate}; the refined loss gains $+5.01$ TEDS-Struc over the replicated baseline on the held-out evaluation set.

The visual foundation in the main text \S\ref{sec:proposed} is only described conceptually; this appendix supplements the complete loss definitions for the Split and Merge stages, in one-to-one correspondence with the code implementation.

\noindent\textbf{Split loss with 1D dilated soft labels.} For pixel-level ground truth $y \in \{0,1\}$, we apply spatial dilation with radius $k=1$ to construct the soft label $y^{soft}[i] = \max_{|j-i| \le k} y[j]$, combined with a binary Focal Loss:
\begin{equation}
    \mathcal{L}_{\text{split}} = w_{\text{row}} \operatorname{FL}(p_r, y_r^{soft}) + w_{\text{col}} \operatorname{FL}(p_c, y_c^{soft})
\end{equation}
where inverse dynamic weighting (IDW) takes $w_{\text{row}} = \mathcal{L}_{\text{col}} / (\mathcal{L}_{\text{row}} + \mathcal{L}_{\text{col}})$, with $w_{\text{col}}$ defined symmetrically. This strategy adaptively compensates for per-axis training accuracy and improves robustness on continuously merged regions across rows/columns.

\noindent\textbf{Merge loss with masked Focal alignment.} The masked Focal Loss for the 4-way OTSL classification is defined as:
\begin{equation}
    \mathcal{L}_{\text{merge}} = \frac{1}{|\Omega|} \sum_{k \in \Omega} \alpha (1 - p_k)^\gamma (-\log p_k)
\end{equation}
where $\Omega$ is the valid-cell set; by setting $\text{ignore\_index} = -100$, the padded region $\Omega^c$ is forcibly excluded, ensuring that the model's predictions of the underlying spatial topology are highly consistent at inference and removing the coordinate-jitter risk of the original scheme in large-scale tensor operations.

\section{Reproducibility Details}
\label{sec:repro}

For reproducibility, we report the decoding hyperparameters of the main inference models (consistent with \texttt{config/api\_config.yaml} in the code repository):

\begin{itemize}
    \item \textbf{LongCat-Flash-Lite} (main-experiment model): temperature 0.0, top\_p 0.95, max\_tokens 20{,}000, seed 33, timeout 600\,s, max\_retries 3.
    \item \textbf{GLM-4.5-Air} and \textbf{GLM-4.6V} (cross-model reference): temperature 0.0, top\_p 0.95, max\_tokens 20{,}000 and 8{,}192 respectively, max\_retries 3.
    \item \textbf{Qwen3-0.6B} (selective routing confirmation and SFT base for the learnable router): default sampling, constrained to output structured JSON labels only.
\end{itemize}

\noindent\textbf{Token accounting.} All token statistics use the native BPE tokenizer of the main inference model, consistent with the efficiency accounting in \S\ref{sec:experimental_setup}.

\noindent\textbf{Inference mode.} All main inference models are invoked via API; locally we only run parsing, routing, and STR rendering; end-to-end latency is significantly affected by network round-trips, so the token-efficiency statistics in this paper focus only on the input-token dimension.

\noindent\textbf{Computing Infrastructure.} The visual parsing models were trained on two NVIDIA RTX 4090 (24GB) GPUs. The split model contains 16.1M parameters and the merge model contains 32.5M parameters. Local inference, including Qwen3-0.6B routing and visual parsing, was executed on a single NVIDIA RTX 4060 GPU. The visual parsing pipeline is highly lightweight, requiring approximately 386\,ms per image and 622\,MB of peak VRAM.

\end{document}